\documentclass[letterpaper, 10 pt, conference]{packages/ieeeconf}  

\IEEEoverridecommandlockouts                              

\usepackage{times}
\usepackage{epsfig}
\usepackage{graphicx}
\usepackage{amsmath}
\usepackage{amssymb}
\usepackage{gensymb}
\usepackage{subfigure} 
\usepackage[ruled]{algorithm}
\usepackage{algpseudocode}
\usepackage{ctable}
\usepackage[export]{adjustbox}
\usepackage{import}
\usepackage{pgf}
\usepackage{url}
\usepackage[bookmarks=true]{hyperref}
\usepackage[dvipsnames]{xcolor}

\usepackage{colortbl}
\makeatletter
\let\NAT@parse\undefined
\makeatother
\usepackage[numbers, sort&compress]{natbib}

\usepackage{packages/sparo_acronyms}
\usepackage{packages/sparo_math}
\let\celsius\relax

\let\degree\relax
\usepackage{packages/sparo_SIunits}
\usepackage{packages/sparo_misc}
\usepackage{multirow}
\usepackage{indentfirst}

\usepackage{soul,color}
\usepackage{verbatim} 

\usepackage{xcolor}
\newcommand{\bl}[1]{{\textcolor{black}{#1}}}

\usepackage[symbol]{footmisc}

\begin{document}

\title{\LARGE \bf
    CUTE-Planner: Confidence-aware Uneven Terrain Exploration Planner

}

\author{Miryeong Park$^{1}$, Dongjin Cho$^{1}$, Sanghyun Kim$^{2}$, and Younggun Cho$^{1\dagger}$
	\thanks{\bl{This work was supported by Institute of Information \& communications Technology Planning \& Evaluation (IITP) grant (RS-2022-II220448), National Research Foundation of Korea (NRF) grant (RS-2025-02217000 and RS-2025-24803365) funded by the Korea government (MSIT), Smart Manufacturing Innovation R\&D funded by Korea Ministry of SMEs and Startups (RS-2024-00448642) and Korea Basic Science Institute (National research Facilities and Equipment Center) grant funded by the Ministry of Science and ICT (No.RS-2025-00564593). }}
	\thanks{$^{1}$Miryeong Park, $^{1}$Dongjin Cho and $^{1\dagger}$Younggun Cho are with the Electrical and Computer Engineering and INHA Future Mobility IPCC, Inha University, Incheon, South Korea 
		{\tt\small [bark9757, d22g66]@inha.edu, yg.cho@inha.ac.kr}}%
	\thanks{$^{2}$Sanghyun Kim is with the Department of Mechanical Engineering, Kyung Hee University, Yongin-si 17104, South Korea
		{\tt\small kim87@khu.ac.kr}}%
}

\maketitle

\begin{abstract} 

Planetary exploration robots must navigate uneven terrain while building reliable maps for space missions. However, most existing methods incorporate traversability constraints but may not handle high uncertainty in elevation estimates near complex features like craters, do not consider exploration strategies for uncertainty reduction, and typically fail to address how elevation uncertainty affects navigation safety and map quality. To address the problems, we propose a framework integrating safe path generation, adaptive confidence updates, and confidence-aware exploration strategies. Using Kalman-based elevation estimation, our approach generates terrain traversability and confidence scores, then incorporates them into Graph-Based exploration Planner (GBP) to prioritize exploration of traversable low-confidence regions. We evaluate our framework through simulated lunar experiments using a novel low-confidence region ratio metric, achieving 69\% uncertainty reduction compared to baseline GBP. In terms of mission success rate, our method achieves 100\% while baseline GBP achieves 0\%, demonstrating improvements in exploration safety and map reliability.

\end{abstract}

\section{Introduction}

Planetary exploration rovers \cite{nasaMoonMissions2024,nasaMarsMissions2024} play an important role in exploring unknown environments and collecting valuable data during space missions. However, teleoperation is limited by communication latency and link interruptions, which hinder real-time hazard response and make remote operations inefficient and unsafe. Accordingly, autonomous navigation capabilities are essential for long-term rover operations.

Mars rovers have demonstrated the critical importance of real-time autonomous navigation without pre-built maps for planetary exploration. Earlier rovers like Spirit and Opportunity \cite{maimone2006mer} used stereo vision systems to build local \ac{DEM} for autonomous navigation. NASA's Perseverance rover \cite{nasaMarsMissions2024} has advanced this capability with the AutoNav system, generating local \ac{DEM} for traversability estimation and onboard path planning, resulting in over 400\,m of additional traverse distance compared with teleoperation. These autonomous navigation techniques demonstrate the importance of real-time \ac{DEM}-based traversability analysis for autonomous exploration, where rovers must simultaneously navigate safely and build environmental maps of unknown planetary terrain.

\begin{figure}[t]
    \centering
    \includegraphics[clip, trim=2mm 2mm 2mm 2mm,
                 height=5.5cm]{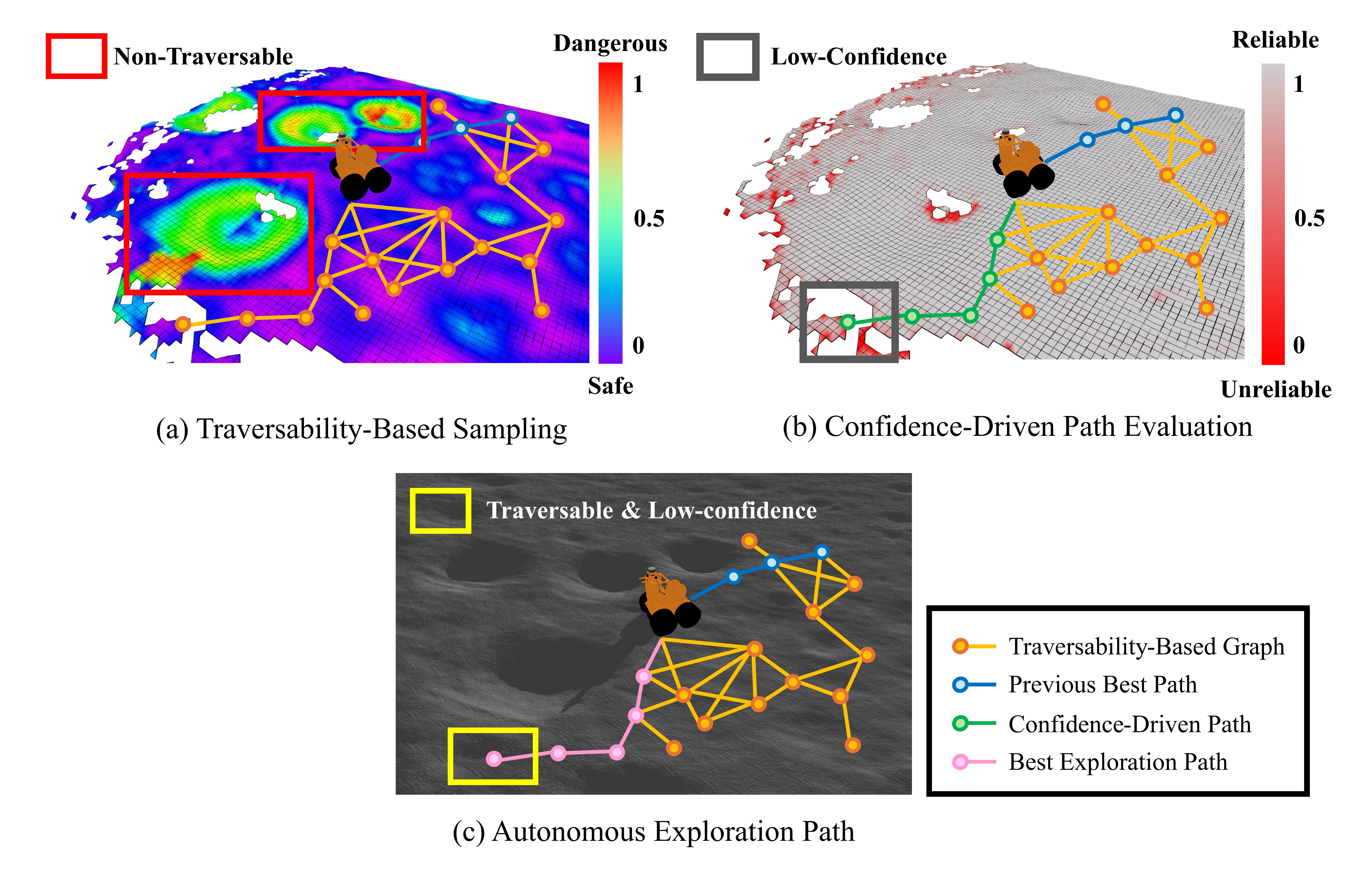}
    \vspace{-0.6cm}
    \caption{
Process of the proposed confidence-aware uneven terrain exploration planner.
(a) Traversability-based sampling: candidate vertices (\textcolor{orange}{orange circles}) are sampled only in traversable regions and connected by \textcolor{orange}{orange lines} to form the traversable local graph in the local traversability map. \textcolor{RoyalBlue}{Blue circles and lines} indicate the previous best exploration path. Lower traversability costs indicate terrain that is easier for robots to navigate.
(b) Confidence-driven path evaluation: the planner selects a path that targets low-confidence regions (\textcolor{darkgray}{gray box}) in the local confidence map; the \textcolor{Green}{green path} represents the confidence-driven exploration path. In the confidence map, lower confidence scores represent higher uncertainty in elevation estimates.
(c) Autonomous exploration path: leveraging both traversability and confidence information produces the best exploration path (\textcolor{Salmon}{pink}), which balances information, safety, and map reliability.
}
    \label{figs:motivation_fig}
    \vspace{-0.4cm}
\end{figure}


For autonomous exploration on planetary surfaces, we identify three key considerations. First, safe and efficient path planning: robots must ensure traversability on uneven terrain for safety while simultaneously selecting informative paths. These requirements are challenging because safe path planning is essential for long-term operation, allowing robots to explore larger areas and acquire more comprehensive terrain data over extended missions. 
Second, uncertainty in terrain models: regions with irregular height variations, such as crater edges, steep inclines, and rock-scattered areas, are difficult to measure due to sensor noise and limited resolution. This results in high-variance estimates, creating low-confidence regions. In such regions, terrain model-based traversability analysis becomes unreliable for safe path planning.
Third, active uncertainty reduction in planning: while existing exploration methods \cite{bi2024lrae, liu2024sfre, patel2025hierarchical} incorporate traversability constraints in path planning, these methods do not address how to reduce uncertainty in regions where traversability estimates are unreliable, leaving high-uncertainty areas in the environmental model. Ignoring this uncertainty can route a rover into hazardous terrain \cite{oh2024trip, tan2025spatiotemporal} and degrade map fidelity for future missions.



To address aformentioned considerations, we propose \textit{CUTE-Planner}, a confidence-aware uneven terrain exploration planner. As illustrated in Fig.~\ref{figs:motivation_fig}, this approach preserves the efficiency of graph-based exploration \cite{dang2020gbp} while explicitly considering for \ac{DEM} uncertainty through real-time evaluation of local confidence and active path planning toward low-confidence regions. In this work, the terrain model is constructed as a \ac{DEM}; we therefore target \ac{DEM} uncertainty explicitly.
Our main contributions are as follows:  
\begin{enumerate}
    \item \textbf{Traversability-based sampling}. To address the challenge of safe and efficient path planning, we ensure that navigation paths are generated through traversable terrain, enabling robots to explore safe regions while maintaining long-term operation for comprehensive terrain mapping.
    
    \item \textbf{Adaptive confidence updates}. We define terrain confidence from the local elevation variance and update it as new sensor data becomes available during exploration, providing real-time uncertainty estimates for navigation planning.
    
    \item \textbf{Exploration with confidence constraints}. Building upon the adaptive confidence updates, we leverage confidence-aware exploration planning by directing the robot toward low-confidence regions. This approach reduces map uncertainty while ensuring safe navigation, improving both immediate path planning reliability and long-term map quality for future missions.
    
    \item \textbf{Low-confidence region ratio for map quality evaluation}. We propose a novel metric that quantifies the reliability of terrain maps generated during exploration. By analyzing confidence value distributions and establishing a statistical threshold, this metric measures the percentage of unreliable terrain data, enabling systematic evaluation of exploration methods based on their ability to produce trustworthy maps rather than merely maximizing coverage.
\end{enumerate}

\begin{figure*}[t]
    \centering
    \def\width{\textwidth}
    {
        \includegraphics[width=\textwidth, clip, trim=0mm 0mm 0mm 40mm]{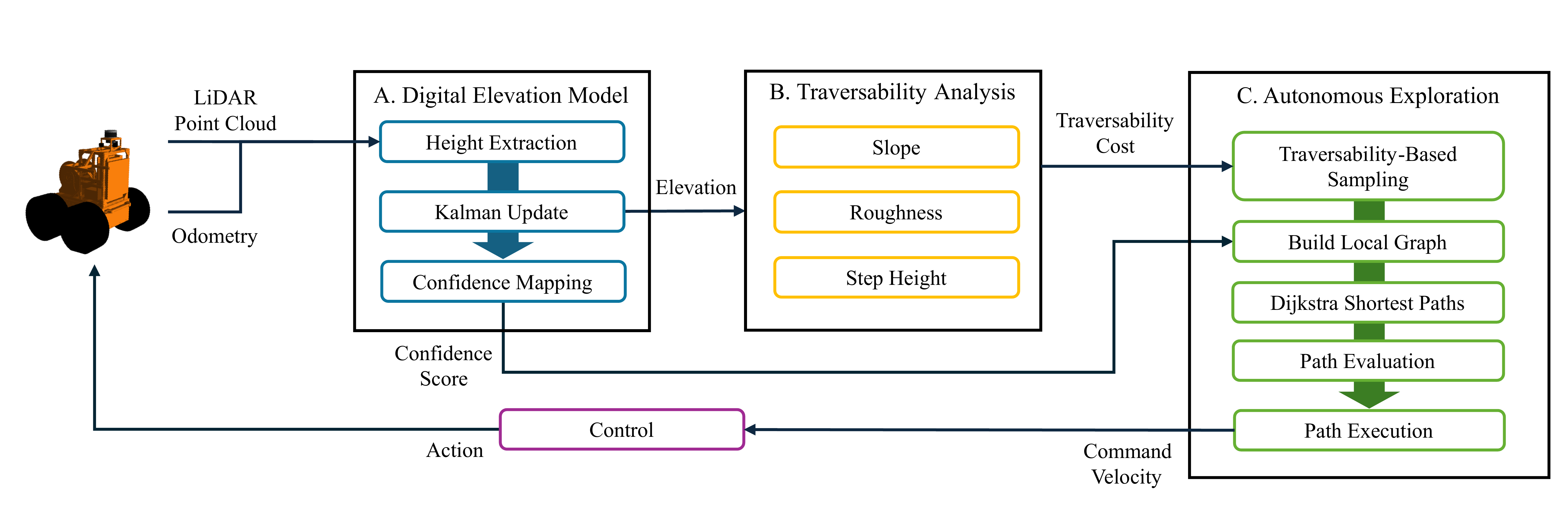}
    }
    \vspace{-0.8cm}
    \caption{Confidence-aware autonomous exploration framework.
A. Kalman estimator for \ac{DEM} construction and confidence mapping,
B. Traversability analysis from terrain attributes (slope, roughness, step height),
C. Autonomous exploration using traversability cost and confidence score.
}
    \vspace{-0.50cm}
    \label{fig:frame_work}
\end{figure*}

\section{Related Work}

\subsection{Terrain Traversability Analysis}
\label{subsection:Traversability}
Traversability analysis is an important task for safe autonomous driving.
Most studies estimate traversability using semantic information from visual sensors, geometric information from depth sensors, or a fusion of both.
\subsubsection{Semantic-based Traversability Analysis}
Several researchers proposed a traversability analysis approach based on predefined costs assigned to semantic classes such as road, grass, and gravel \cite{hosseinpoor2021traversability,roth2310viplanner}.
In contrast, \citet{guan2022ga} proposed a method that directly predicts the traversability cost for each segment from RGB images. 
However, these approaches require large amounts of training data and remain dependent on predefined cost categories. To address these problems, self-supervised approaches \cite{kahn2021badgr, aegidius2025watch} have been introduced to reduce dependence on labeled data and predefined cost categories. However, since these methods still utilize semantic information, terrain attributes such as slope, roughness, and height are relatively less emphasized.
\subsubsection{Geometry-based Traversability Analysis}
Various studies have utilized \ac{DEM} to estimate terrain attributes. A \ac{DEM} is a 2.5D representation in which each cell of a 2D grid stores the elevation value at the corresponding location \cite{fankhauser2018probabilistic, werner2024covariance, cremean2005uncertainty, forster2015continuous, schoppmann2021multi}. 
Based on \ac{DEM}, terrain attributes such as slope, roughness, and step height can be computed, and these serve as key metrics in traversability analysis \cite{chilian2009stereo, shan2018bayesian, leininger2024gaussian}. In this work, \ac{DEM}-based traversability mapping is adopted due to the inherent lack of semantic information in planetary environments.

\subsection{Autonomous Exploration}
Autonomous exploration research has traditionally been divided into two types of methods—frontier-based and sampling-based planning—while more recent work has explicitly incorporated traversability awareness.
\subsubsection{Frontier-Based Exploration}
The frontier concept guided a robot toward the boundary between known free and unknown space \cite{yamauchi1997frontier}.
Later studies \cite{batinovic2021multi, dai2020fast} reduced computation with multi-resolution detection and voxel grouping.
Because these planners followed a greedy strategy, they could miss globally optimal viewpoints.
\subsubsection{Sampling-Based Exploration}
The \ac{NBVP} \cite{bircher2016receding} grew a \ac{RRT} and followed the branch with maximum gain.
\ac{GBP} \cite{dang2020gbp} clustered a local \ac{RRG} with \ac{DTW} to construct a sparse global graph. The approach proved its exploration efficiency by achieving first place in the \ac{DARPA} Subterranean Challenge \cite{DARPA_SubT_2018}.
The TARE \cite{cao2021tare} method formulated viewpoint visiting as an extended \ac{TSP}, and \citet{huang2023fael} ranked graph edges sampled at a uniform spatial resolution according to exploration gain.
\subsubsection{Traversability-Aware Exploration}
\citet{patel2024stage} coupled a semantic traversability with a graph-based exploration planner, while a hierarchical terrain-aware framework \cite{patel2025hierarchical} blended geometric and semantic costs for multimodal robots.
\citet{liu2024sfre} embedded geometric costs in an exploration module. Similarly, \citet{bi2024lrae} employed a sampling-based planner guided by large-region heuristics while enforcing slope and roughness constraints.
These approaches passively incorporated traversability costs into exploration planning, potentially facing difficulties when traversability estimates were unreliable.
In contrast, the proposed framework actively corrects misestimated traversability while maintaining the efficiency of a graph-based exploration.

\section{Methodology}

In this paper, our proposed autonomous exploration framework's pipeline is shown in \figref{fig:frame_work}. First, a local elevation map is constructed from \ac{LiDAR} point clouds. Second, the system evaluates terrain traversability on this elevation map while concurrently computing a confidence score for each cell. Finally, the autonomous exploration module samples candidate vertices based on traversability cost and utilizes cell confidence for path evaluation. This process is repeated until the environment has been fully explored or the robot's battery capacity is exhausted.

\subsection{Digital Elevation Model}
In this paper, elevation and confidence are calculated using \ac{LiDAR} point clouds. These values are continuously updated using a Kalman estimator \cite{welch1995introduction} to construct a real-time local elevation map in the form of a 2.5D grid map. First, the height of each point cloud is obtained by transforming and projecting them into the map frame. The resulting heights are stored in an \(\mathrm{N}\times \mathrm{N}\) grid map \(\mathbf{G}^{(i,j)}\), where \((i,j)\) denotes the cell coordinates. For each grid cell, the elevation $h$ and its variance $\sigma^2$ are updated online by a scalar Kalman filter processing the latest \ac{LiDAR} points.

\paragraph*{Measurement Model}
Following Fankhauser et al.~\cite{fankhauser2018probabilistic}, we define the measurement model in this section. We consider only range measurement noise and ignore pose errors, since the evaluation is performed in simulation where the ground-truth robot pose is known exactly and thus introduces no additional uncertainty. In the map frame $M$, height measurements can be represented by Gaussian probability distribution $\tilde{z}_k \sim \mathcal{N}(\bar{z}_k,\sigma_\mathrm{meas}^2)$. Here, $\bar{z}_k$ is the mean, and $\sigma_\mathrm{meas}^2$ is the variance.

In the map frame $M$, the height scalar $\bar{z}_k$ is as follows:
\begin{equation}
  \bar{z}_k = \mathbf{P}_z\bigl(\mathbf{\Phi}_{MS}(r_{P}^{S}) + r_{M}^{S}\bigr),
\end{equation}
where \(\mathbf{P}_z = [0\;0\;1]\) is the projection function that extracts the \(z\)-component, \(\mathbf{\Phi}_{MS}\in SO(3)\) denotes the rotation matrix from the sensor frame \(S\) to the map frame \(M\). \(r_{P}^{S}\in\mathbb{R}^3\) denotes the position of measurement \(P\) in \(S\), and \(r_{M}^{S}\in\mathbb{R}^3\) represents the map origin expressed in \(S\).

In the measurement model, the measurement variance is given as follows:
\begin{equation}
  \sigma_{\mathrm{meas}}^2 = \mathbf{J}_S \,\boldsymbol{\Sigma}_S\, \mathbf{J}_S^\top,
\end{equation}
where \(\boldsymbol{\Sigma}_S\) is the covariance of the sensor’s range noise. This variance accounts for both the effect of sensor tilt, captured by the Jacobian \(\mathbf{J}_S\), and the sensor’s noise. It is then incorporated into the Kalman gain to update the estimate and its associated uncertainty.

The Jacobian of the height scalar \(\bar{z}_k\) with respect to the sensor frame point \(r_{P}^{S}\) is given by
\begin{equation}
    \mathbf{J}_S \;=\; \frac{\partial \bar{z}_k}{\partial r_{{P}}^{S}}.
\end{equation}
The Jacobian matrix \(\mathbf{J}_S\) provides a linear approximation of how perturbations in the sensor frame point along the \(x\)-, \(y\)-, and \(z\)-axes affect the height \(\bar{z}_k\).

\paragraph*{Kalman Prediction}
The elevation $h_k^{(i,j)}$ at time $k$ and cell $(i,j)$ is represented by a first-order state space model.
We assume there is no error in the robot pose, and since the terrain is static, the state transition parameter is set to be $a=1$ and the process noise set to be $Q_{k-1}=0$.
The prediction value is calculated from the previous estimation as follows:
\begin{align}
\hat{h}_{k|k-1}^{(i,j)}   &= a\,\hat{h}_{k-1|k-1}^{(i,j)},\\   
{\sigma^2}_{k|k-1}^{(i,j)}&= a^{2}{\sigma^2}_{k-1|k-1}^{(i,j)}+Q_{k-1},
\end{align}
\noindent where ${\sigma^2}_{k|k-1}^{(i,j)}$ is the prior variance, which represents the uncertainty of the elevation estimate before new measurements are obtained. For the initial time step ($k=0$), the measurement values are used directly as the initial estimates.

\paragraph*{Kalman Update}
Given the prior estimate $(\hat{h}_{k|k-1}^{(i,j)}, {\sigma^2}_{k|k-1}^{(i,j)})$ and the new measurement $(\bar{z}_k, \sigma_{\text{meas}}^{2})$, the Kalman gain is defined as:

\begin{equation}
\label{eq:Kalman_gain}
K_{k}=\frac{{\sigma^2}_{k|k-1}^{(i,j)}}{{\sigma^2}_{k|k-1}^{(i,j)}+\sigma_{\text{meas}}^{2}}.
\end{equation}

\noindent Based on~\eqref{eq:Kalman_gain}, we can update the elevation and variance as follows:
\begin{align}
\label{eq:height_update}
\hat{h}_{k|k}^{(i,j)}&=\hat{h}_{k|k-1}^{(i,j)}
+K_{k}\bigl(\bar{z}_k-\hat{h}_{k|k-1}^{(i,j)}\bigr),\\
{\sigma^2}_{k|k}^{(i,j)}&=(1-K_{k}){\sigma^2}_{k|k-1}^{(i,j)}.
\label{eq:variance_update}
\end{align}

\paragraph*{Confidence Mapping}
The posterior variance ${\sigma^2}_{k|k}^{(i,j)}$ from~\eqref{eq:variance_update} directly measures elevation uncertainty. We convert this into a normalized confidence score as follows:
\begin{equation}
\label{eq:confidence}
  \mathcal{C}_k^{(i,j)} = 1 - \operatorname{clip}\,\!\bigl(\sigma_{k|k}^{2};\,0,\,1\bigr),
  \quad \mathcal{C}_k^{(i,j)} \in [0,1],
\end{equation}
\noindent where \(\operatorname{clip}\,(x;0,1)\) limits \(x\) to \([0,1]\). 
Thus, \(\mathcal{C}=1\) indicates maximum confidence, while \(\sigma_{k|k}^{2}\!\ge\!1\) yields \(\mathcal{C}=0\).
This definition reflects both sensor noise and observation frequency: as the number of updates increases, the posterior variance \(\sigma_{k|k}^2\) decreases.

\subsection{Traversability Analysis}
To estimate terrain traversability, we evaluate three geometric attributes—slope, roughness, and step height. 
Each attribute is computed from a neighborhood matrix \(\mathbf{N}^{(i,j)} \in \mathbb{R}^{(2n+1)\times(2n+1)}\) centered at \(\mathbf{G}^{(i,j)}\). 
Its elements are the Kalman-filtered elevations of \eqref{eq:height_update} as follows:
\begin{equation}
\mathbf{N}^{(i,j)} =
\begin{bmatrix}
\hat{h}_{k|k}^{(i-n,j-n)} & \cdots & \hat{h}_{k|k}^{(i-n,j+n)} \\
\vdots & \hat{h}_{k|k}^{(i,j)} & \vdots \\
\hat{h}_{k|k}^{(i+n,j-n)} & \cdots & \hat{h}_{k|k}^{(i+n,j+n)}
\end{bmatrix},
\label{eq:n_matrix}
\end{equation}

\subsubsection{Slope Estimation}

The slope at \( \mathbf{G}^{(i,j)} \) is estimated from \( \mathbf{N}^{(i,j)} \). Given that \( \mathbf{N}^{(i,j)}_{n,n} = \hat{h}_{k|k}^{(i,j)} \), we construct a set of 3D points by mapping each element \( \mathbf{N}^{(i,j)}_{m,l} \) in the window to:
\begin{equation}
\mathbf{p}_{m,l} = 
\begin{bmatrix} 
\ (i + m - n) \cdot \delta \ \ \\
\ (j + l - n) \cdot \delta \ \ \\
\mathbf{N}^{(i,j)}_{m,l}
\end{bmatrix},
\label{eq:point3d}
\end{equation}

\noindent where \(\delta\) denotes the grid resolution and \( m, l \in \{0, \dots, 2n\} \). The covariance matrix of the resulting point set is computed as follows:
\begin{equation}
\mathbf{Cov}^{(i,j)} = \frac{1}{(2n+1)^2} \sum_{m=0}^{2n} \sum_{l=0}^{2n} (\mathbf{p}_{m,l} - \bar{\mathbf{p}})(\mathbf{p}_{m,l} - \bar{\mathbf{p}})^\top,
\end{equation}
\noindent where \( \bar{\mathbf{p}} \) denotes the mean of all points \( \mathbf{p}_{m,l} \) in the window. The surface normal is estimated via \ac{PCA} as the eigenvector of \( \mathbf{Cov}^{(i,j)} \) associated with the smallest eigenvalue \( \lambda_3 \), assuming the eigenvalues are ordered as \( \lambda_1 \geq \lambda_2 \geq \lambda_3 \).
The local slope angle \( s^{(i,j)} \) is computed as the angle between the eigenvector \( \mathbf{v}_3 \) corresponding to the smallest eigenvalue \( \lambda_3 \) and the vertical axis \( \mathbf{e}_z = [0,\; 0,\; 1]^\top \) as follows:
\begin{equation}
s^{(i,j)} = \cos^{-1}\left(|\mathbf{v}_3 \cdot \mathbf{e}_z|\right) \cdot \frac{180}{\pi},
\end{equation}

\subsubsection{Roughness Estimation}

The roughness \( r^{(i,j)} \) measures local terrain unevenness as follows:
\begin{equation}
r^{(i,j)} = \frac{1}{(2n+1)^2 - 1} 
\sum_{m=0}^{2n} \sum_{l=0}^{2n}
\left| \mathbf{N}^{(i,j)}_{m,l} - \mathbf{N}^{(i,j)}_{n,n} \right|,
\end{equation}
\noindent where \( \mathbf{N}^{(i,j)}_{n,n} \) corresponds to the center cell of \( \mathbf{N}^{(i,j)} \).

\subsubsection{Step Height Estimation}
The step height \( d^{(i,j)} \) represents the maximum local terrain elevation difference as follows:
\begin{equation}
  d^{(i,j)} = 
    \max_{m,l}\,
    \lvert \mathbf{N}^{(i,j)}_{m,l} -
            \mathbf{N}^{(i,j)}_{n,n} \rvert,  
\end{equation}

\subsubsection{Traversability Cost}

The traversability cost \( T^{(i,j)} \) is defined as a weighted sum of normalized slope, roughness, and step height:
\begin{equation}
\label{eq:traversability_cost}
T^{(i,j)} = \omega_1 \frac{s^{(i,j)}}{s_{\text{crit}}} 
+ \omega_2 \frac{r^{(i,j)}}{r_{\text{crit}}} 
+ \omega_3 \frac{d^{(i,j)}}{d_{\text{crit}}},
\end{equation}
where \( \omega_1, \omega_2, \omega_3 \in [0,1] \) are weighting coefficients totaling 1. 
The terms \( s_{\text{crit}}, r_{\text{crit}}, d_{\text{crit}} \) denote robot-specific thresholds for slope, roughness, and step height.  
Lower values of \( T^{(i,j)} \) correspond to more traversable terrain.

\subsection{Confidence-Aware Autonomous Exploration}

This section presents the integration of (i) traversability-based sampling, (ii) adaptive confidence updates for local graph, and (iii) exploration with confidence constraints into the baseline \ac{GBP} framework \cite{dang2020gbp}.

\subsubsection{Traversability-Based Sampling}
\label{subsec:traversability_sampling}

Given a random sample $\xi_\mathrm{rand}=[x_r,y_r]^\top$ in the map frame,
the corresponding grid indices are
\begin{equation}
\label{eq:grid_coordinates}
  i_r = \Bigl\lfloor \tfrac{x_r}{\delta} \Bigr\rfloor,\qquad
  j_r = \Bigl\lfloor \tfrac{y_r}{\delta} \Bigr\rfloor,
\end{equation}
where $\delta$ is the grid resolution, and $\lfloor\cdot\rfloor$ is the floor function that maps a real number to the largest integer less than or equal to it.
We assume that the grid coordinate system shares the same origin and axis directions as the map coordinate system. With robot radius $r_{\mathrm{rob}}$,
the maximum index offset inside the robot's circular footprint is
\begin{equation}
\label{eq:frob}
  F_\mathrm{rob} = \Bigl\lceil \tfrac{r_{\mathrm{rob}}}{\delta} \Bigr\rceil .
\end{equation}
Here, $\lceil\cdot\rceil$ is the ceiling function that maps a real number to the smallest integer greater than or equal to it.
\noindent Define the offset set
\begin{equation}
\label{eq:offset_set}
  \mathcal{O} =
  \Bigl\{\,(\Delta i,\Delta j)\in\mathbb{Z}^2 \,\bigl|\,
         \Delta i^{2} + \Delta j^{2} \le F_\mathrm{rob}^{2} \Bigr\}.
\end{equation}
Each $(\Delta i,\Delta j)$ represents a cell inside the circular footprint
centered at $(i_r,j_r)$.
The sample $\xi_\mathrm{rand}$ is accepted as a graph vertex $v$
only if
\begin{equation}
\label{eq:traversability_check}
  T^{(i_r + \Delta i,\; j_r + \Delta j)}
  \;\le\; T_{\max}
  \quad
  \forall\,(\Delta i,\Delta j) \in \mathcal{O},
\end{equation}
where $T^{(i_r + \Delta i,\; j_r + \Delta j)}$ is the traversability cost from~\eqref{eq:traversability_cost}
and $T_{\max}$ is a user-defined threshold.
If any cell violates \eqref{eq:traversability_check}, the candidate is
rejected and resampled.

\subsubsection{Adaptive confidence updates for local graph}

Let the local graph \(\mathcal{G}_L\) be defined as
\begin{equation}
\mathcal{G}_L = (\mathcal{V}, \mathcal{E}, \mathcal{C}),
\end{equation}
where $\mathcal{V}$ is the set of vertices, $\mathcal{E} \subseteq \mathcal{V} \times \mathcal{V}$ is the set of edges, and $\mathcal{C}$ is the set of a confidence score with each vertices.
For a vertex $v \in \mathcal{V}$ at position $(x_v,y_v)$ in the map coordinates, the corresponding cell coordinates $(i_v,j_v)$ are determined by~\eqref{eq:grid_coordinates}. The vertex confidence is initially set as:
\begin{align}
  \mathcal{C}_v &= \mathcal{C}^{(i_v,j_v)},
\end{align}
where $\mathcal{C}^{(i_v,j_v)}$ is from~\eqref{eq:confidence}. As new sensor observations update the \ac{DEM}, vertex confidence values are adaptively updated:
\begin{equation}
  \mathcal{C}_v \leftarrow \max\bigl(\mathcal{C}_v,\;\mathcal{C}^{\mathrm{new}}_{v}\bigr),
\end{equation}
where $\mathcal{C}^{\mathrm{new}}_{v}$ denotes the confidence newly estimated at vertex $v$ from the current \ac{DEM} update. This ensures that each vertex maintains the highest confidence observed for its corresponding cell location.


\subsubsection{Exploration with confidence constraints}
\label{subsec:confidence_gain}

The confidence-based component of the exploration gain for a candidate path \(\gamma = \{v_0,\ldots,v_n\}\) is defined as
\begin{equation}\label{eq:confidence_gain}
  g_{\mathrm{conf}}(\gamma)=
  \max_{v_l\in\gamma}
  \begin{cases}
    1, & \!\!\!\!\!\mathcal{C}_{v_l} \ge \Theta_\mathcal{C},\\[4pt]
    \exp\!\bigl(\beta\,(\Theta_\mathcal{C} - \mathcal{C}_{v_l})\bigr), &\!\!\!\!\! \mathcal{C}_{v_l} < \Theta_\mathcal{C},
  \end{cases}
\end{equation}

\noindent where \(\mathcal{C}_{v_l}\) is the confidence score at vertex \(v_l\), \(\Theta_\mathcal{C}\in[0,1]\) is a user-defined confidence threshold, and \(\beta>0\) controls the sensitivity of the confidence gain. Increasing $\beta$ steepens the exponential term for vertices with $\mathcal{C}_{v_l} < \Theta_\mathcal{C}$, so paths that traverse low-confidence regions receive proportionally larger $g_{\text{conf}}$ values, and the confidence term carries greater relative weight in the overall exploration gain. If all vertices within the path have confidence values above the confidence threshold, the confidence gain becomes 1.
The overall exploration gain is then obtained by
\begin{equation}
\label{eq:exploration_gain}
  g(\gamma) = g_{\mathrm{vol}}(\gamma) \times g_{\mathrm{conf}}(\gamma),
\end{equation}
where \(g_{\mathrm{vol}}(\gamma)\) is the \ac{GBP} volumetric gain of the path. If there is even one low-confidence vertex within the path, the confidence gain and volumetric gain are multiplied.

\algrenewcommand\algorithmicrequire{\textbf{Input:}}

\begin{algorithm}[t]
  \caption{Local Exploration}
  \label{alg:local_exploration}
  \begin{algorithmic}[1]
    \Require Current robot state $\xi_0$, Local traversability map $M^T$, Traversability threshold $T_{\max}$, Local confidence map $M^\mathcal{C}$, Confidence threshold $\Theta_\mathcal{C}$
    \State $\xi_0 \gets \textsc{GetCurrentRobotState}()$
    \State $\xi_{\text{rand}} \gets \textsc{SampleTraversableVertex}(M^T, T_{\max})$ 
    \State $\xi_{\text{nearest}} \gets \textsc{NearestNeighborSearch}(\xi_0)$ 
    \State $\mathcal{G}_L \gets \textsc{BuildLocalGraph}(\xi_0,\;\xi_{\text{rand}},\;\xi_{\text{nearest}}, M^\mathcal{C})$ 
    \State $\Gamma_L \gets \textsc{GetDijkstraShortestPaths}(\mathcal{G}_L,\;\xi_0)$
    \State \textsc{ComputeVolumetricGain}($\mathcal{G}_L$)
    \State $g_{\text{best}} \gets 0$ \quad $\Gamma_{L,\text{best}} \gets \varnothing$
    \ForAll{$\gamma \in \Gamma_L$}
      \State $g_{\text{vol}} \gets \textsc{ExplorationGain}(\gamma)$
      \State $g_{\text{conf}} \gets \textsc{ComputeConfidenceGain}(\gamma,\;\Theta_\mathcal{C})$
      \State $g_\gamma \gets g_{\text{vol}} \times g_{\text{conf}}$
      \If{$g_\gamma > g_{\text{best}}$}
        \State $g_{\text{best}} \gets g_\gamma$
        \State $\Gamma_{L,\text{best}} \gets \gamma$
      \EndIf
    \EndFor
    \State $\Gamma_{L,\text{best}} \gets \textsc{OptimizePath}(\Gamma_{L,\text{best}})$ 
    \State \Return $\Gamma_{L,\text{best}}$
  \end{algorithmic}
\end{algorithm}

\subsubsection{Local Exploration}
\label{subsec:local_exploration}

The overall local exploration process is shown in the Algorithm~\ref{alg:local_exploration}. At each planning iteration, the local planner builds a small graph \(\mathcal{G}_L\) around the robot using the local traversability map \(T_{\mathrm{local}}\) and the current robot state \(\xi_0\) (3D position and heading). The procedure consists of four stages:

\begin{enumerate}
  \item \textbf{Traversability-Based Sampling.}  
    As detailed in Sec.~\ref{subsec:traversability_sampling}, each random sample \(\xi_{\mathrm{rand}}\) is drawn within \(T_{\mathrm{local}}\) but is only accepted if all grid cells beneath the robot’s circular footprint satisfy the traversability threshold. Samples failing this check are discarded and retried, guaranteeing traversability safety.

  \item \textbf{Build Local Graph.}  
    \begin{itemize}
      \item Find the closest existing node \(\xi_{\mathrm{nearest}}\in \mathcal{G}_L\) to the sample \(\xi_{\mathrm{rand}}\) using Euclidean distance.
      \item If the straight‐line path between \(\xi_{\mathrm{nearest}}\) and \(\xi_{\mathrm{rand}}\) is traversable, add \(\xi_{\mathrm{rand}}\) as a new vertex and connect it to \(\xi_{\mathrm{nearest}}\).
      \item For the new vertex, attempt to connect to all neighbors within a predefined radius under collision‐free conditions.
      \item Repeat sampling and connection until \(\mathcal{G}_L\) contains a sufficient number of vertices.
    \end{itemize}
  \item \textbf{Shortest Path Search.}  
    Run Dijkstra’s algorithm on \(\mathcal{G}_L\) to compute the set of shortest paths \(\Gamma_L\) from the current node to all other nodes.

  \item \textbf{Exploration Gain Evaluation \& Execution.}  
    For each \(\gamma\in\Gamma_L\), compute~\eqref{eq:exploration_gain} and select
    \(\displaystyle \gamma^* = \arg\max g(\gamma)\),
    then send the first segment of \(\gamma^*\) to the path execution module. After execution, the planner repeats this process, continually guiding the robot toward areas with high volumetric gain and low-confidence regions.  
\end{enumerate}

\section{Experimental Evaluation}\label{sec:experiments}

We benchmark the proposed planner against \ac{GBP} \cite{dang2020gbp} and
\textit{Only\_Trav} in two Gazebo lunar environments, \textit{Moon~1} and \textit{Moon~2} \cite{HybridAstarMPPI}.

\subsection{Experimental Setup}\label{subsec:experimental_setup}

\begin{figure}[!t]
    \centering
	\def\width{\columnwidth}%
	\includegraphics[width=\width, clip, trim=0mm 4mm 0mm 0mm]{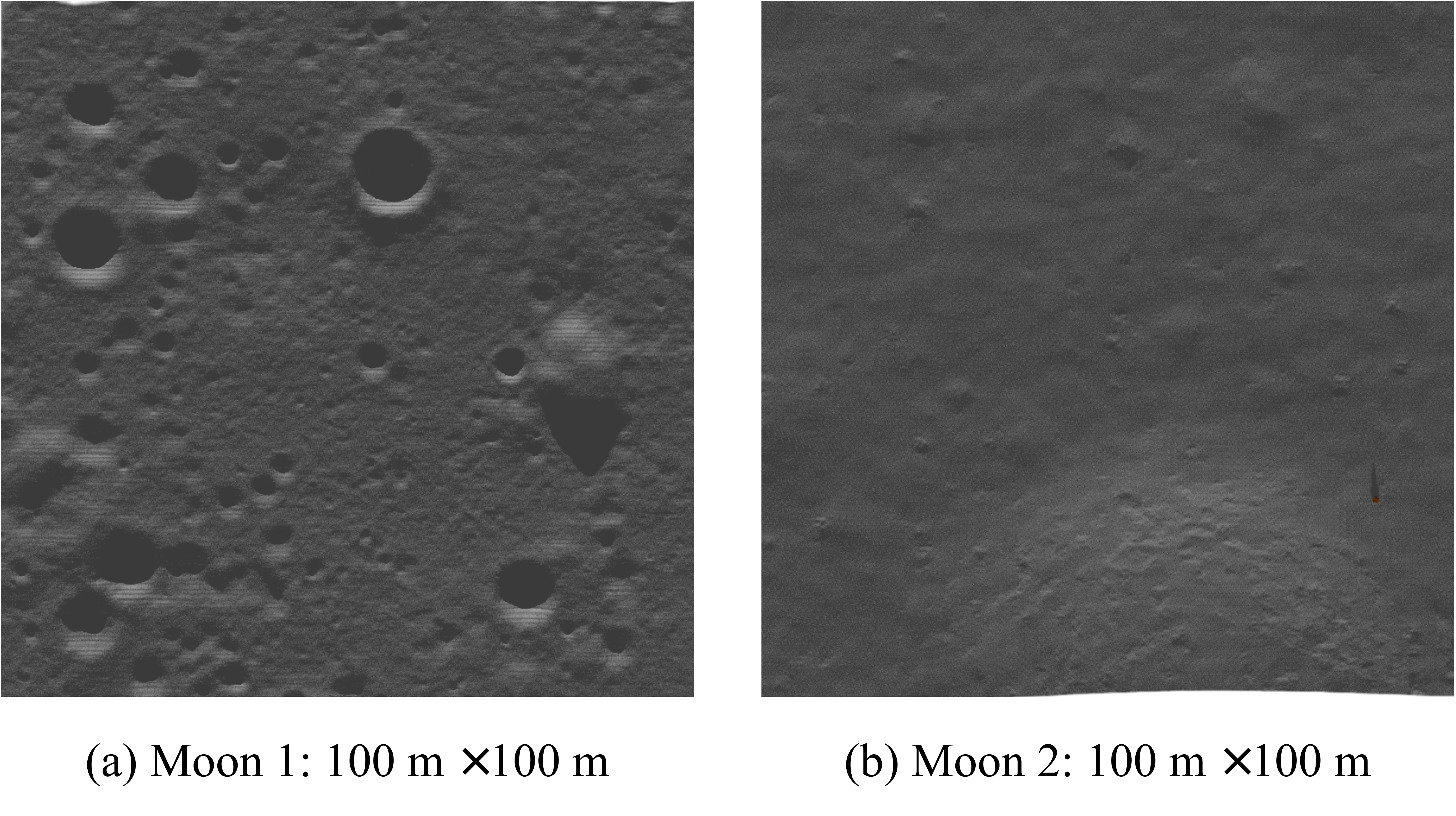}
    \vspace{-0.8cm}
 \caption{Lunar simulation environments: (a) \textit{Moon~1}, (b) \textit{Moon~2}.}
    \label{figs:environment}
  \vspace{-0.5cm}
\end{figure}

All experiments were conducted using the ETH RobotX SuperMegaBot equipped with a Velodyne VLP-16 LiDAR operating at 10\,Hz. Traversability thresholds were set to $s_{\text{crit}} = 20\degree$, $r_{\text{crit}} = 0.15\,$m, $d_{\text{crit}} = 0.2\,$m, and $T_{\max} = 0.4$, with a confidence threshold of $\Theta_\mathcal{C} = 0.8$. Trials took place in two Gazebo lunar arenas, each measuring \(100\,\mathrm{m} \times 100\,\mathrm{m}\), as shown in \figref{figs:environment}. 
A linear battery model was employed to represent the \ac{SoC} dynamics during operation. Let $C$ [Ah] denote the nominal battery capacity and $C_0$ [Ah] represent the initial available capacity, with $C = C_0$. The SoC at time $t$ is defined as:
\begin{equation}
    \mathrm{SoC}(t) = \left(\frac{C_0}{C} - \frac{1}{C}\int_0^t \frac{I(\tau)}{3600}\,d\tau\right) \times 100\%
\end{equation}
where $I(\tau)$ denotes the instantaneous current [A] at time $\tau$ [s]. The discharge current was assumed constant at $I_0 = 1.44C$ throughout each run, corresponding to a linear SoC decrease of $0.04\%$ per second. Each run was terminated after 2400\,s (40\,min) to maintain a $4\%$ safety margin, preventing over-discharge.
We compared our planner with \textit{Only\_Trav}, which uses the same traversability-based sampling but evaluates paths using volumetric gain only, and with \ac{GBP}, which relies on random sampling and volumetric gain without traversability filtering or confidence updates. Each planner was executed five times per environment from randomized initial poses (consistent across methods for fair comparison). All other parameters remained unchanged across methods.

\subsection{Exploration Rate}
\begin{figure}[!t]
    \centering
	\def\width{\columnwidth}%
	\includegraphics[width=0.98\width, clip, trim=0mm 4mm 0mm 4mm]{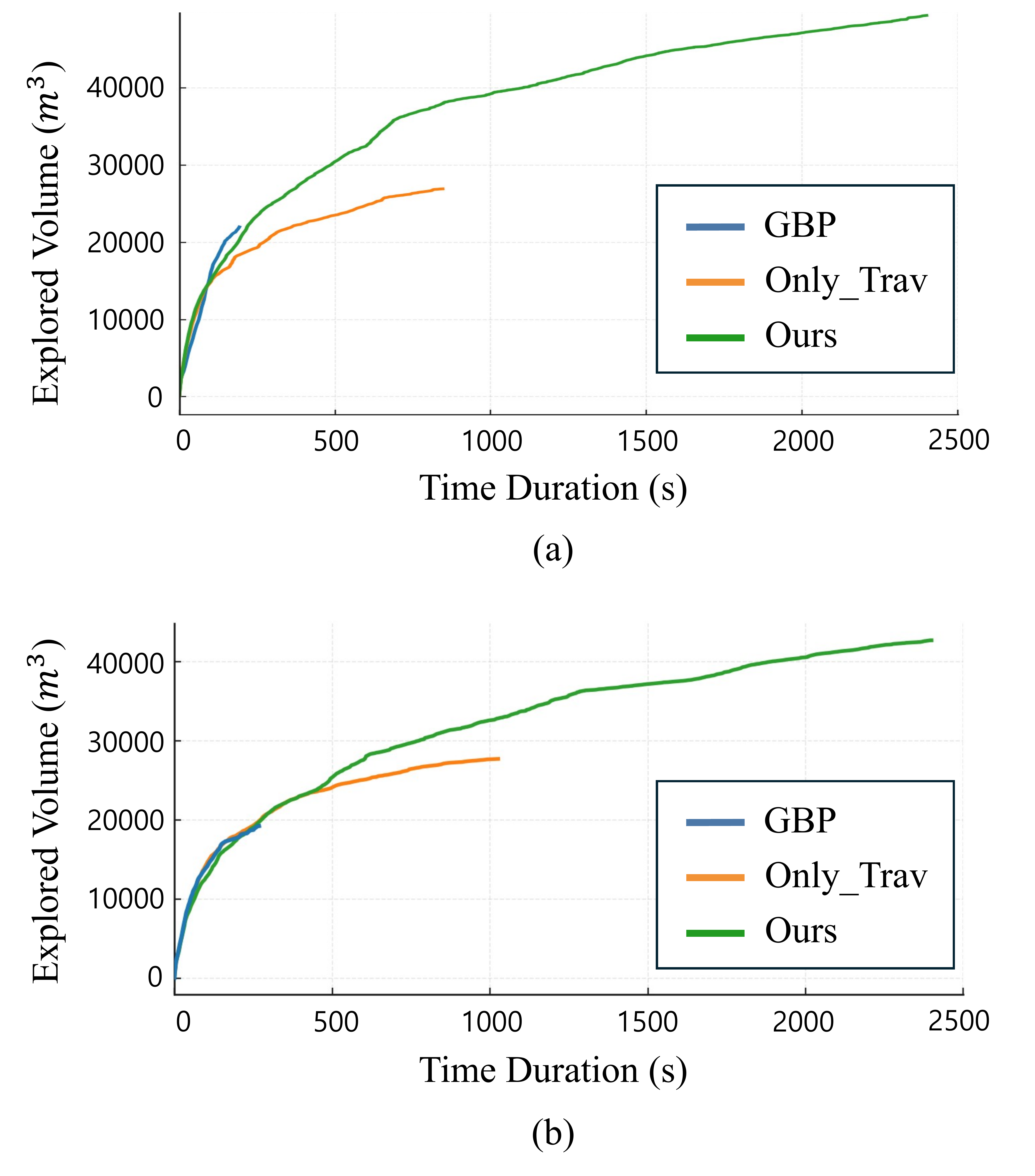}
    \vspace{-0.4cm}
 \caption{Average explored volume over time in \textit{Moon~1} and \textit{Moon~2}.}
    \label{figs:exploration_rate}
  \vspace{-0.4cm}
\end{figure}

\begin{figure}[!t]
    \centering
	\def\width{\columnwidth}%
	\includegraphics[width=\width, clip]{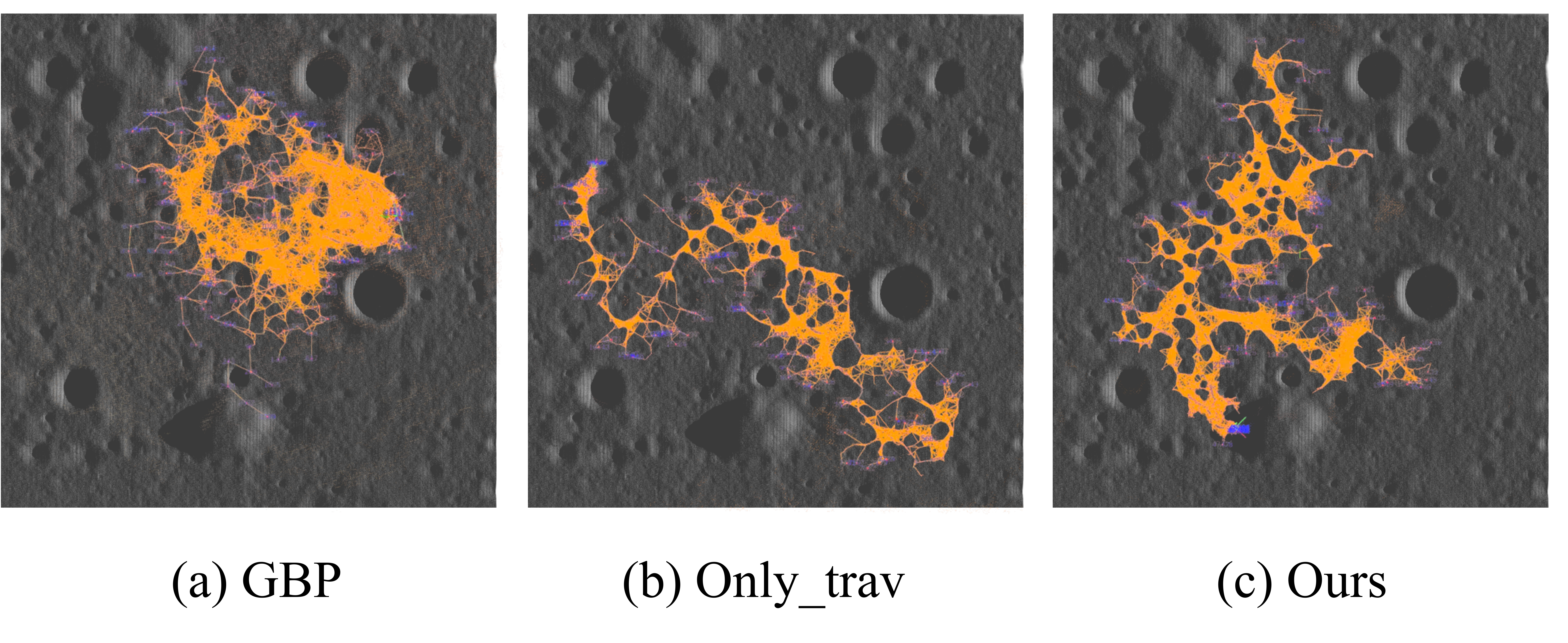}
    \vspace{-0.8cm}
 \caption{Qualitative comparison of exploration 
 results in \textit{Moon~1}. \textcolor{orange}{Orange graph} shows the planning graph generated during exploration. (a) GBP, (b) Only\_Trav, (c) Ours.}
    \label{figs:graph_result}
  \vspace{-0.6cm}
\end{figure}

\figref{figs:exploration_rate} compares the cumulative exploration volume over time in both lunar simulation environments. The graph plots the average values obtained from five experiments conducted in each environment. The proposed method achieves the largest final exploration volume in both \textit{Moon~1} and \textit{Moon~2}. The baseline method \ac{GBP} explores terrain marginally faster during the initial exploration phase in \textit{Moon~1}, but its mission terminates early due to safety constraints, resulting in limited total exploration coverage. The \textit{Only\_Trav} approach sustains exploration longer than \ac{GBP}, but ultimately encounters difficulties maintaining long-term operation due to unreliable traversability estimates. Qualitative comparisons of the planning graphs are shown in~\figref{figs:graph_result}.

\subsection{Mission Success Rate and Time Duration}
\label{subsec:missiontime}
A mission is considered successful if the robot completes the full 40\,min operation without tipping over. With five trials in each environment, the overall success rate is reported out of ten total trials. The average operating time reflects early terminations due to navigation failures. Table~\ref{tab:mission_overall} shows that only the proposed planner remained upright throughout every trial and therefore accumulated the full 2400\,s of operating time. \textit{Only\_Trav} mitigates some hazards but still fails to complete most trials. The baseline GBP did not successfully complete a single mission, and its average operating time was only approximately 9.6\% of that achieved by the proposed method. These results demonstrate that the proposed approach significantly enhances driving safety compared to other methods.

\begin{table}[H]
  \centering
  \caption{Mission Success Rate and Average Time Duration}
  \label{tab:mission_overall}
  \begin{tabular}{lcc}
    \toprule
    \textbf{Method} & \textbf{Success Rate (N/10)} & \textbf{Avg.\ Time (s)} \\
    \midrule
    GBP         &  0/10 & 230.7  \\
    Only\_Trav  &  2/10 & 936.1 \\
    Ours        &  \textbf{10/10} & \textbf{2400} \\
    \bottomrule
  \end{tabular}
\end{table}

\subsection{Low-Confidence Region Ratio}
We discretize all observed local confidence values into bins of width 0.1 and count their frequencies over 40\,min of operation. Defining the top 95\% as \textit{high-confidence} and the remaining 5\% as \textit{low-confidence}, we find that cells with confidence \(\le 0.8\) account for 5.02\% of observations. Therefore, we set the confidence threshold in~\eqref{eq:confidence_gain} to \(\Theta_C=0.8\). To assess performance at the map level, we accumulate confidence values into a global confidence map for each run. We then compute the percentage of cells in this global map with confidence \(\le0.8\), which serves as our low-confidence region ratio. A lower low-confidence region ratio indicates that the elevation map is more reliable, reflecting more trustworthy traversability estimates. As summarized in Table~\ref{tab:lcratio}, the proposed planner reduces
this ratio by approximately 36\% relative to \textit{Only\_Trav}
and by about 69\% relative to \ac{GBP}, nearly halving the amount of unreliable terrain data produced by the baseline.

\begin{table}[H]
  \centering
  \caption{Low-Confidence Region Ratio (\%)}
  \label{tab:lcratio}
  \begin{tabular}{lcc}
    \toprule
    \textbf{Method} & \textit{Moon~1}\, & \textit{Moon~2}\,\ \\
    \midrule
    GBP         & 5.27 & 3.03 \\
    Only\_Trav  & 2.37          & 1.65        \\
    Ours        & \textbf{1.47}  & \textbf{1.10}      \\
    \bottomrule
  \end{tabular}
\end{table}

\section{CONCLUSION}

This paper presents a confidence-aware exploration framework for autonomous planetary robots operating in uneven terrain. The proposed method integrates Kalman-based elevation estimation, geometric traversability analysis, and confidence-driven path planning within the \ac{GBP} framework. Extensive simulations in lunar terrain environments demonstrate that the approach significantly improves map reliability by reducing low-confidence regions while achieving higher total coverage and consistent mission success. The results show that incorporating uncertainty awareness into exploration planning leads to more robust autonomous exploration in challenging planetary environments. While this study assumed ground-truth robot poses in simulation to isolate the effects of elevation uncertainty, real-world deployments will inevitably encounter localization errors. Future work will address robustness under localization uncertainty, validate the planner in real-world scenarios, and develop advanced uncertainty models for complex terrain.

\footnotesize	
\bibliographystyle{IEEEtranN} 
\bibliography{packages/string-short, packages/references}

\end{document}